\documentclass[11pt]{article}

\usepackage[margin=1in]{geometry}

\usepackage{amsmath,amssymb}

\usepackage{graphicx}
\usepackage{epstopdf} % optional; fine to keep if you use EPS

\usepackage[table]{xcolor}
\usepackage{array}

\usepackage[colorlinks=true,linkcolor=blue,citecolor=blue,urlcolor=blue]{hyperref}

\usepackage[numbers,sort&compress]{natbib}
\usepackage{microtype}
\begin{document}
\vspace*{0.2in}

\begin{flushleft}
{\Large
\textbf\newline{VillageNet: Graph-based, Easily-interpretable, Unsupervised Clustering for Broad Biomedical Applications} % Please use "sentence case" for title and headings (capitalize only the first word in a title (or heading), the first word in a subtitle (or subheading), and any proper nouns).
}
\newline
\\
Aditya Ballal\textsuperscript{1},
Gregory A. DePaul\textsuperscript{2},
Esha Datta\textsuperscript{3},
Asuka Hatano\textsuperscript{1,4},
Erik Carlsson\textsuperscript{2},
Ye Chen-Izu\textsuperscript{1,5,6},
Javier E. L\'{o}pez\textsuperscript{6,7,8},
Leighton T. Izu\textsuperscript{1*},
\\
\bigskip
\textbf{1} Department of Pharmacology, University of California, Davis, California, United States
\\
\textbf{2} Department of Mathematics, University of California, Davis, California, United States
\\
\textbf{3} Sandia National Laboratories, 
 Albuquerque, NM, 87108, USA
\\
\textbf{4} Mechanical Engineering Department, the University of Tokyo, Tokyo, Japan
\\
\textbf{5} Biomedical Engineering, University of California, Davis, California, United States
\\
\textbf{6} Internal Medicine, University of California, Davis, California, United States
\\
\textbf{7} Cardiovascular Medicine, University of California, Davis, California, United States
\\
\textbf{8} Cardiovascular Research Institute, University of California, Davis, California, United States
\\
\bigskip
* ltizu@ucdavis.edu

\end{flushleft}
\newpage

\section*{Abstract}
Clustering complex, large-scale biomedical data is essential for precision medicine applications. Because biomedical data may reveal latent biological patterns or subgroups with significant clinical outcomes, clustering these data is important for downstream tailoring of medical therapies for distinctive patient subgroups. Complexity in biomedical data may originate from inherently variable features of datasets and/or heterogeneous sources of information, such as electronic health records or physiological, cellular, and/or molecular assays. The more novel and/or complex a biomedical dataset is, the less is usually known at the offset about its inherent features, e.g. labels, linearity, etc. that can limit the initial selection of suitable clustering techniques.

Building upon our previous work (i.e., MapperPlus), we introduce VillageNet, an unsupervised clustering framework that integrates topological principles, graph-based community detection, and random-walk analysis to derive data-driven knowledge in an unsupervised context. VillageNet autonomously infers the number of clusters directly from the data and demonstrates a robust ability to identify clusters with non-linear separation, thereby avoiding restrictive assumptions about cluster geometry, a commonly unknown feature of biomedical datasets.

VillageNet was evaluated on an extensive suite of non-biomedical benchmark datasets with known ground-truth labels, as well as four heterogeneous biomedical datasets (flow cytometry, tissue imaging, single-cell gene expression, and image-derived data). VillageNet achieved overall superior performance when assessed using normalized mutual information and an adjusted Rand index, and favorable computational properties, with runtime scaling linearly with both dataset size and dimensionality—thereby eliminating the need for dimension-reduction procedures. Together, these findings establish VillageNet as a scalable, topology-informed, and broadly generalizable framework for clustering complex biomedical datasets, especially during the discovery phase when most features about complex datasets may still be unknown.

\section*{Author summary}
Modern biomedical research generates complex, high-dimensional datasets that span patient-level records, cellular measurements, and tissue-scale imaging. These datasets often lack suitable training sets and furthermore, variability across patients, experiments, and/or platforms can make labels learned in one setting unreliable in another. Moreover, the open-ended nature of many biomedical problems makes it difficult to know in advance what patterns should be expected or what labels should be used to train analysis pipelines. For this reason, unsupervised clustering plays a central role in biomedical data analysis by enabling the discovery of unknown groups directly from the data. Currently, many existing clustering approaches are tailored to specific data types or rely on assumptions that do not always generalize across modalities. VillageNet addresses this gap by integrating ideas from topological data analysis (TDA) with graph partitioning, enabling robust and scalable clustering of diverse biomedical settings, including single-cell gene expression, flow cytometry, and tissue imaging.

\section{Introduction}

Unsupervised clustering is a fundamental machine-learning (ML) technique for identifying hidden or latent patterns in data without requiring prior labels. This is particularly valuable in biomedical research, where there is often a need to group samples based on shared characteristics, yet annotations are unavailable or unreliable due to inherent biological and/or experimental variability\cite{biomedicineclustering}. In flow cytometry, for example, marker expression levels for a given cell type can differ substantially between experiments, so cells are typically clustered purely from the marker expression profiles within that experiment; biological identities are assigned afterward based on the characteristics of each cluster\cite{flow_unsupervised}. A similar challenge arises in single-cell RNA sequencing, where gene expression varies widely across individuals and batches; here, unsupervised clustering of expression profiles serves as the first step before interpreting cell types\cite{SS_unsupervised}. On a broader clinical scale, clustering of electronic health records (EHRs)—which integrate local demographic variables, laboratory values, imaging outputs, and clinical histories—can reveal clinically significant patient subgroups that can guide diagnosis, prognosis, and personalized treatment strategies. By uncovering biologically and/or clinically significant groupings, unsupervised clustering can help improve disease understanding, inform new drug development, and drive more targeted and perhaps more effective healthcare interventions.

Biomedical datasets tend to be high-dimensional, heterogeneous in structure and source, noisy, and sometimes characterized by nonlinear relationships that are not always known beforehand\cite{nonlinearbio}. Available clustering algorithms such as K-Means\cite{KMeans}, fuzzy C-Means\cite{CMeans}, and K-Medoids\cite{KMedoid} assume spherical structure or linear separability within datasets and require knowing the number of clusters to perform—assumptions and information rarely met in biological systems. Density-based algorithms such as DBSCAN\cite{DBSCAN} and OPTICS\cite{OPTICS} can detect unknown clusters on data with arbitrary shapes but are strongly dependent on density hyperparameters that are sensitive to noise commonly found in biological measurements. 

Graph-based clustering algorithms, including Louvain\cite{Louvain}, Leiden\cite{Leiden}, and Phenograph\cite{Phenograph}, have become widely used in single-cell and imaging analyzes due to their scalability and empirical effectiveness\cite{Scanpy}. These methods operate on nearest-neighbor graphs, which provide a powerful way to capture local structure but can be challenging to construct reliably in very high-dimensional or heterogeneous datasets. In addition to these methods, kernel-based clustering techniques represent another class of approaches that map data into higher-dimensional feature spaces to capture complex relationships \cite{Kernelreview}. Kernel extensions of K-Means \cite{KernelKmeans} and fuzzy C-Means \cite{KernelCMeans} have been proposed to improve clustering performance in such settings, but often incur an increased computational cost and require careful selection of kernel functions. In general, these methods represent powerful tools, but each carries assumptions or design choices that may be difficult to satisfy across a full spectrum of biomedical data. 

To address these challenges, we developed VillageNet, a clustering algorithm that integrates concepts from topological data analysis (TDA), network theory, and random-walk–based community detection. Inspired by our previous work MapperPlus \cite{MapperPlus}, VillageNet reduces the complexity of high-dimensional data by constructing a structured multi-scale representation that captures both local and global organization. We evaluated VillageNet across a broad and heterogeneous suite of publicly available non-biomedical benchmark datasets and find that its performance is consistently competitive with or superior to other state-of-the-art clustering approaches. To explore its relevance in precision medicine, we further apply VillageNet to four heterogeneous biomedical datasets spanning single-cell RNA sequencing, flow cytometric per-cell analysis, tissue imaging, and image-derived cellular profiling. Across all these modalities, VillageNet yielded biologically interpretable clusters, highlighting its suitability for a wide range of biomedical applications.

\section{Method}

\subsection{Motivation}

VillageNet is based on the assumption that although cluster boundaries may be globally non-linear, they can be regarded as locally linear when examined at sufficiently small scales. This principle underlies several approaches in topological data analysis (TDA), including the Mapper algorithm \cite{Mapper}, on which our previous work MapperPlus\cite{MapperPlus} is based, that decomposes complex datasets into smaller regions where local structure can be more easily characterized.

In Mapper, this decomposition is achieved by partitioning the data space into overlapping bins along selected \textit{lens} functions, followed by clustering within each bin. Although effective in certain settings, this approach is highly sensitive to hyperparameter and lens-function choices, scales poorly with dimensionality, and may result in a substantial fraction of data points being labeled as noise.

VillageNet adopts a different strategy to identify local linear regions. Rather than discretizing the feature space uniformly, we deliberately overcluster the dataset using K-Means, partitioning the data into $\nu$ fine-grained partial clusters, which we refer to as \textit{villages}. Each  village corresponds to the Voronoi cell associated with its K-Means centroid, so that every data point within a village is closer to its centroid than to any other. By choosing $\nu$ sufficiently large, each Voronoi cell becomes small enough that the local structure of the cluster boundaries around it can be effectively approximated as linear. These villages serve as coarse-grained units for subsequent analysis.

\subsection{Relationship between Villages}

After coarse-graining the dataset into villages, the next goal is to reveal the global structure by analyzing the boundary between each pair of villages. In this context, for a data point $i$ in village $V$ and a different village $U$, we define
\begin{equation}
\mathcal{D}(\overrightarrow X_i,U) = \min \left\{ \|\overrightarrow y\|: \|\overrightarrow X_i - \overrightarrow y - \overrightarrow \mu_U\| = \|\overrightarrow X_i - \overrightarrow y - \overrightarrow \mu_{V}\|\right\},
\end{equation}  
where $\|.\| $ denotes the $l_2$ distance and $\overrightarrow \mu_U$ and $\overrightarrow \mu_{V_i}$ denote the centroids of the villages $U$ and $V$, respectively. Intuitively, $\mathcal{D}(\overrightarrow X_i,U)$ measures the distance between the data point $\overrightarrow X_i$ and the boundary between the villages $U$ and $V$. This distance can also be expressed as a projection along the line that connects the centroids of the two villages.
\begin{equation}
\mathcal{D}(\overrightarrow X_i,U) = \left( \overrightarrow{X}_i - \frac{1}{2}( \overrightarrow{\mu}_U + \overrightarrow{\mu}_{V_i} )\right) \cdot \frac{\overrightarrow{\mu}_{V}-\overrightarrow{\mu}_U}{\|\overrightarrow{\mu}_{V}-\overrightarrow{\mu}_U\|}.
\end{equation}  
\textbf{Exterior of Village:} Using this boundary-based distance, we define the \textit{exterior} of a village $U$ as the set of data points that lie close to its boundary with neighboring villages:
\begin{equation}
U^{(E)}= \{i \mid \mathcal{D}(\overrightarrow X_i,U) < \epsilon_U \text{ and } i \notin U \},
\end{equation}  
where $\epsilon_U$ is a parameter that controls the neighborhood size of the village $U$. To ensure a consistent neighborhood size across all villages, for each village $U$, the parameter $\epsilon_U$ is chosen such that $|U^{(E)}| = \eta$.  The hyperparameter $\eta$ is conceptually similar to the perplexity parameter in t-SNE. This condition is satisfied by partially sorting the distances $[\mathcal{D}(\overrightarrow X_i,U) \ \forall i \notin U]$ to select the smallest $\eta$ value. 

\textbf{Inter-Village Graph:} The exteriors provide a natural mechanism to quantify connectivity between villages. We construct a weighted graph, referred to as the \textit{village network}, in which each node represents a village. The weight of the edge between villages $U$ and $V$ is defined as
\begin{equation}
A_{UV} = |U^{(E)} \cap V| + |U \cap V^{(E)}|.
\end{equation} 
This definition quantifies the number of data points that lie near the shared boundary between the two villages, accounting for proximity from both sides. The resulting adjacency matrix $A$ is symmetric. Larger edge weights indicate a high density of data points along the boundary between villages, reflecting strong connectivity, whereas smaller weights indicate weaker connections.

\subsection{Community Detection on the Village Network:}

Once the \textit{village network} is constructed, the next step is to identify groups of villages that are strongly connected to each other. To achieve this, we employ the walk-likelihood community finder (WLCF) \cite{WLCF} that is based on the principle that random walkers initialized within a community are more likely to remain within a community than to move outside it. The algorithm combines maximum likelihood estimates of node visit frequencies obtained from random walks with modularity optimization to accurately partition the graph. This graph-clustering step is conceptually similar to MapperPlus\cite{MapperPlus}, where a coarse-grained graph representation of the data is likewise partitioned to reveal global structure. A key advantage of this approach is that the optimal number of communities naturally emerges from the algorithm, rather than being specified as an input.

\textbf{Final Clustering:} To form the final clusters, we simply merge all data points from the villages that belong to the same community. Since no data point belongs to more than one village, the final clusters are also non-overlapping.

\subsection{Algorithm}

\begin{figure}[h]
\begin{center}
%\framebox[4.0in]{$\;$}
%\fbox{\rule[-.5cm]{0cm}{4cm} \rule[-.5cm]{4cm}{0cm}}
\includegraphics[scale=0.5]{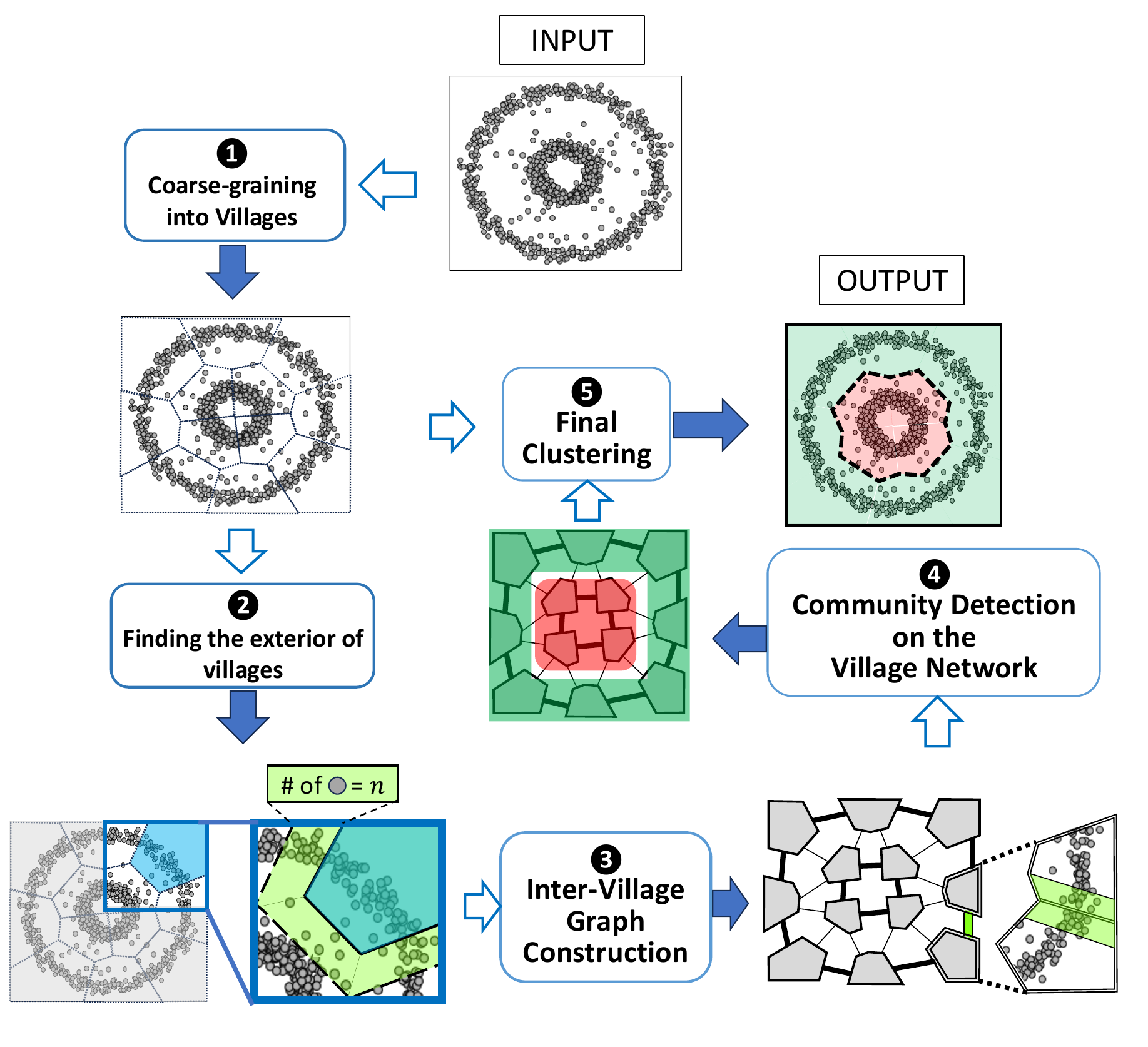}
\end{center}
\caption{Sequential Stages of VillageNet Clustering }
\label{fig:methoddiagram}
\end{figure}

Thus, the VillageNet clustering algorithm can be summarized in the following five steps, as shown in Fig \ref{fig:methoddiagram}
\begin{enumerate}
    \item \textbf{Coarse-graining into Villages:} The dataset is subdivided using K-Means into $\nu$ partial-clusters or $\nu$ \textit{villages}. Each instance $i$ is assigned a unique village $V$.
    \item  \textbf{Finding the exterior of villages:} For each village $U$, we define its exterior $U^{(E)}$ as the $\eta$ data points outside $U$ that are nearest to the boundary between $U$ and their own village.
    \item \textbf{Inter-Village Graph Construction:} Based on the identified exteriors, a weighted graph—termed the \textit{village network}—is constructed. Each node corresponds to a village, and the edge weight between villages $U$ and $V$ is computed from the number of exterior points of $U$ assigned to $V$, together with the number of exterior points of $V$ assigned to $U$.
    \item \textbf{Community Detection on the Village Network:} The village network is then divided into communities using WLCF, which automatically determines the number of communities.
    \item \textbf{Merging Villages:} All data points belonging to villages within the same community are merged to form the final clusters.
\end{enumerate}

\subsection{Algorithm Complexity}

The time complexity of each step in our algorithm is as follows:
\begin{enumerate}
    \item \textbf{Coarse-graining into Villages:}  This step constitutes the most computationally intensive component of the algorithm. While the K-Means problem is NP-hard in general \cite{KMeanslimitation}, VillageNet requires only a locally optimal solution rather than a global optimum. Accordingly, we employ Lloyd’s algorithm\cite{Lloyd_1982} with a single random initialization, using a customized configuration of the \texttt{scikit-learn} K-Means \cite{sklearn} implementation instead of its default settings. Under these choices, the computational complexity of this step is $O(N*\nu*d)$, where $N$ represents the number of datapoints, $\nu$ is the number of villages, and $d$ is the dimensionality of the dataset. 
    \item   \textbf{Finding the exterior of villages:} The most computationally expensive task in this step is the partial sorting of lists, which is performed using the \textit{numpy argpartition} function \cite{numpy}. The algorithmic complexity of partial sorting is $O(N)$, and this operation is carried out $\nu$ times, once for each village. Thus, the overall time complexity of this step is $O(N*\nu)$.
\item \textbf{Inter-Village Graph Construction:} Building the village network involves multiplying a $\nu \times N$ matrix by an $N \times \nu$ matrix. This step has a nominal complexity of $O(N*\nu^2)$, but in practice, optimized \textit{numpy} matrix operations make it significantly faster than the above clustering steps \cite{numpy, matmulcomplexity}.
\item  \textbf{Community Detection on the Village Network:} Empirical observations suggest that the time complexity of WLCF is approximately $O(\nu^{1.5})$, where $\nu$ represents the number of nodes in the network \cite{WLCF}.
\item \textbf{Merging Villages:} This step is the least computationally intensive and only requires $O(N)$ time.

\end{enumerate}

\begin{figure}[h]
\begin{center}
%\framebox[4.0in]{$\;$}
%\fbox{\rule[-.5cm]{0cm}{4cm} \rule[-.5cm]{4cm}{0cm}}
\includegraphics[scale=0.1]{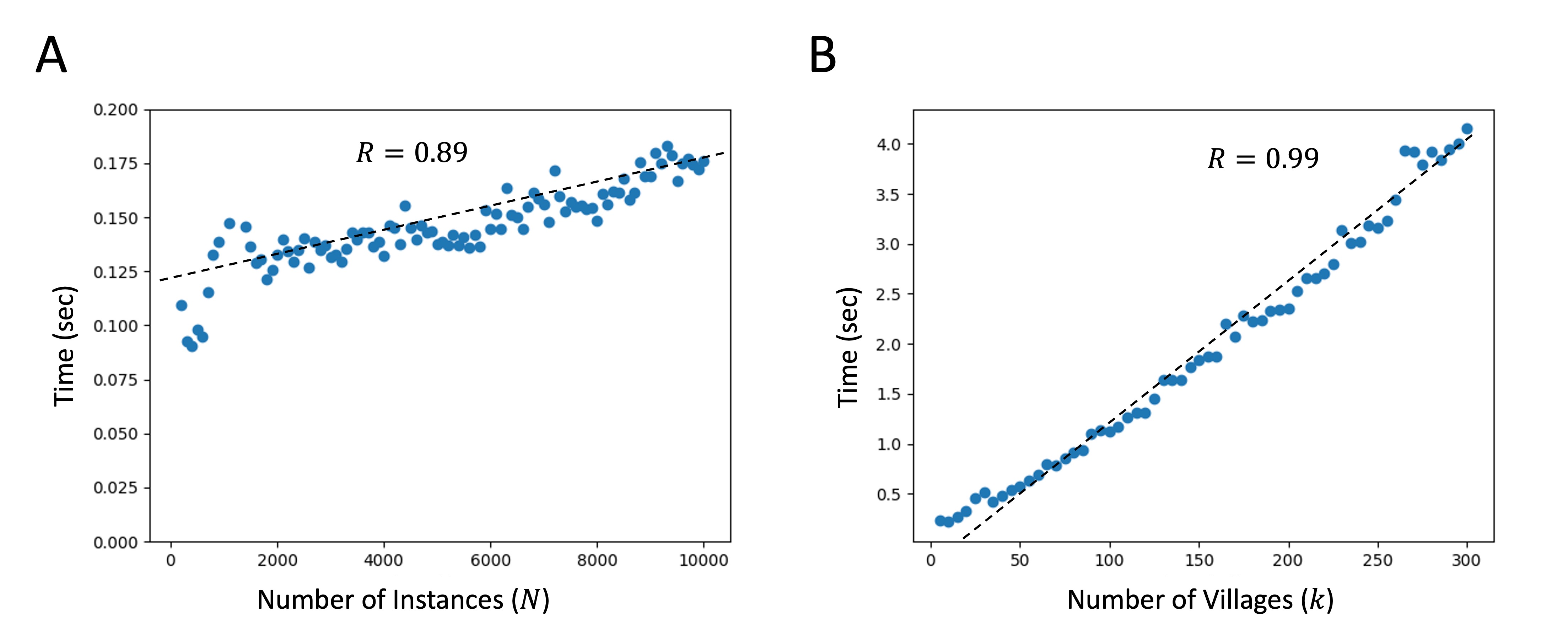}
\end{center}
\caption{Wall Time Analysis of VillageNet Clustering on Various Implementations of the Two-Moons Dataset}
\label{fig:timecomplexity}
\end{figure}

Thus, in the limit of $\nu\ll N$, the overall time complexity of the algorithm approaches $O(N*\nu*d)$. To validate this, Fig \ref{fig:timecomplexity} displays the wall time of VillageNet on different implementations of the two moons dataset. Fig \ref{fig:timecomplexity}A illustrates variations in the size of the dataset with a constant $\eta=50$, while Fig \ref{fig:timecomplexity}B shows the varying numbers of communities for a fixed size of the dataset. Empirically, the observed time complexity aligns with the theoretically predicted one, as depicted in Fig \ref{fig:timecomplexity}.

\subsection{Effect of hyperparameters}

\begin{figure}[h]
\begin{center}
%\framebox[4.0in]{$\;$}
%\fbox{\rule[-.5cm]{0cm}{4cm} \rule[-.5cm]{4cm}{0cm}}
\includegraphics[scale=0.6]{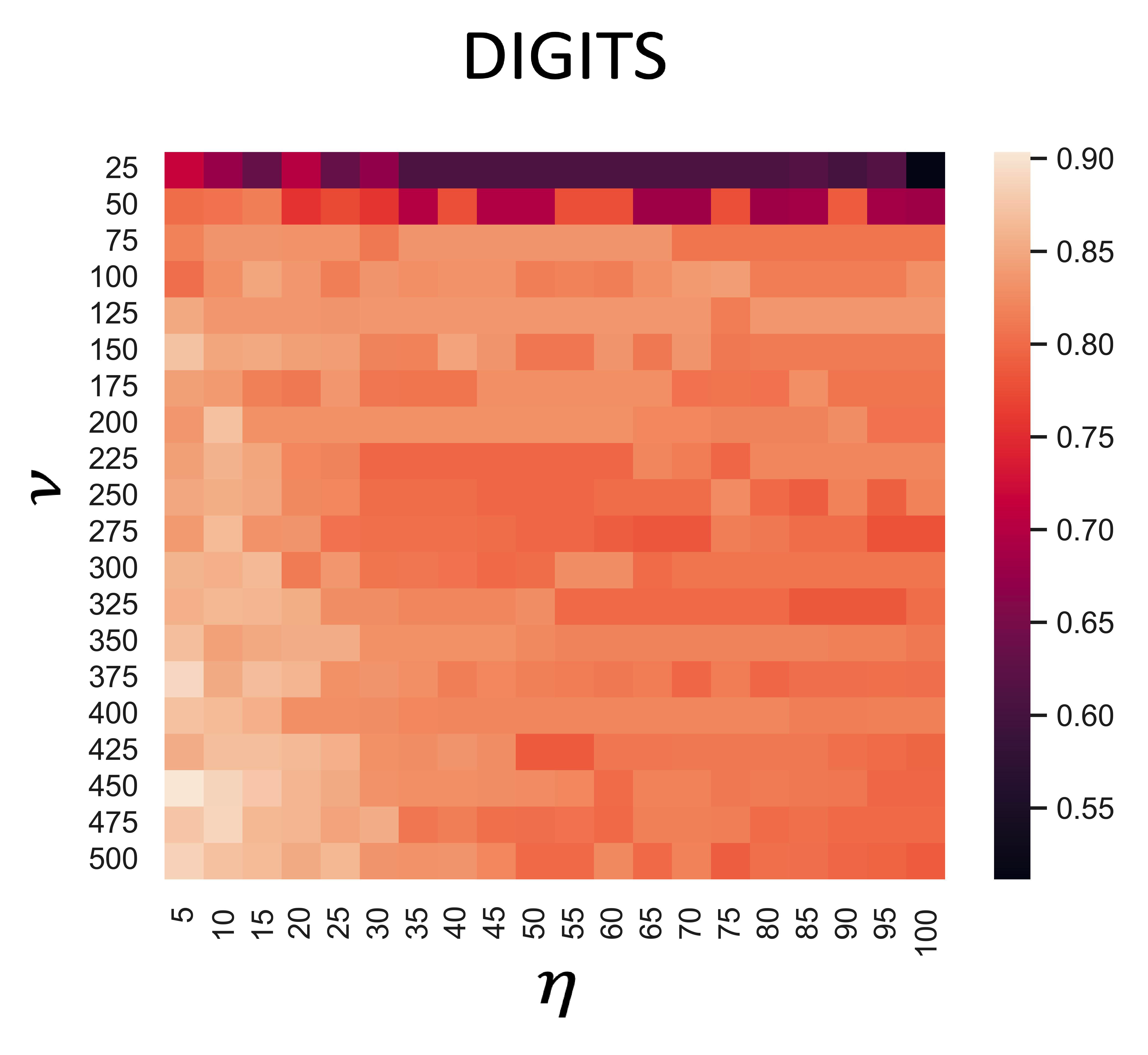}
\end{center}
\caption{Comparison of clusters obtained by VillageNet Clustering on different hyperparameters with the ground truth on Digits dataset}
\label{fig:stability}
\end{figure}

Fig \ref{fig:stability} provides insights into the impact of VillageNet hyperparameters on the final clustering results, evaluated on the digits dataset. The assessment is based on the comparison of the obtained clusters with the ground truth, measured by Normalized Mutual Information (NMI). This step of the algorithm involves two crucial hyperparameters:

\begin{enumerate}
    \item $\nu$: The performance of our algorithm is closely related to the choice of parameter $\nu$. Since, in step 4 of our algorithm, each data point within a village is assigned the same cluster identity as the village itself, the accuracy of our algorithm in clustering the dataset is contingent on how effectively K-Means clustering can shatter the true labels of the dataset. This accuracy tends to improve with larger values of $\nu$ as observed in Fig \ref{fig:stability}. However, as $\nu$ increases, so does the algorithm's runtime linearly.

       Therefore, selecting the optimal value for $\nu$ involves a trade-off. On the one hand, a higher $\nu$ may improve the accuracy of the clustering; on the other hand, it increases computational time. Thus, the ideal choice for $\nu$ lies in finding the minimum number of K-Means clusters that can approximately represent the underlying structure of the data.
    \item  $\eta$: This hyperparameter $\eta$ plays a critical role in shaping connectivity within the village graph. In the extreme scenario where $\eta$ equals $N$, each individual data point becomes a village, essentially transforming the village network into a graph of the nearest neighbor $\eta$ of the dataset. In contrast, when $\eta<N$, the village network provides a coarser representation of the nearest neighbor $\eta$ in the graph.

The value of $\eta$ is pivotal in expressing the connectivity between villages. It should be chosen judiciously, considering the balance between granularity and coarseness in representing local structures. If $\eta$ is too large, it can lead to a dilution of information on the local structures within the villages, as observed in Fig \ref{fig:stability}. This dilution poses a challenge in obtaining correct clusters, emphasizing the importance of appropriately tuning $\eta$ to capture the desired level of connectivity without sacrificing the fidelity of local information within villages.

\end{enumerate}

\subsection{Software Availability}

The software is publicly available at \href{https://github.com/lordareicgnon/VillageNet}{https://github.com/lordareicgnon/VillageNet} and at \href{https://villagenet.streamlit.app}{https://villagenet.streamlit.app}.

\section{Experiments}

\subsection{Datasets}

We evaluated the performance of VillageNet Clustering on a diverse suite of heterogeneous datasets, including eight publicly available benchmark datasets \cite{UCI} and four biomedical datasets including Breast Cancer\cite{BCdata}, Pollen Single-Cell transcriptomic dataset \cite{Pollen}, Levine 32 flow-cytometry dataset \cite{Levine} and ORGANAMNIST dataset \cite{OrganAMNIST}. Each dataset has known ground-truth cluster labels, enabling quantitative evaluation of clustering accuracy.  

Table~\ref{tab:datasetinformation} summarizes the datasets, listing the number of samples, attributes,ground-truth clusters, and type of data. The collection spans a wide range of domains, sizes, and data modalities—from low-dimensional tabular data to high-dimensional image embeddings and biomedical measurements. This diversity enables a robust assessment of the performance of VillageNet in varying clustering challenges.

\begin{table}[!ht]
\centering
\caption{\bf Summary of publicly available datasets used for evaluation. Ground-truth labels are available for all datasets to enable quantitative comparison.}

%\resizebox{\textwidth}{!}{%
\begin{tabular}{|l|l|l|l|l|}
\hline
Dataset & Observations & Attributes & Known Clusters & Data Type \\ \hline
Pen Digits       & 1,797  & 64   & 10  & Images \\ \hline
MNIST            & 70,000 & 784  & 10  & Images \\ \hline
FMNIST           & 70,000 & 784  & 10  & Images \\\hline
 Gas Sensor& 13910& 130& 6&Sensor readings\\\hline 
UCI-HAR& 10299& 561& 6& Heterogenous Variables\\ \hline
Dry Beans        & 13,611 & 16   & 7& Image description\\ \hline
Wine             & 178    & 13   & 3   & Heterogeneous Variables \\ \hline
WiFi             & 2,000  & 7    & 4   & Heterogeneous Variables \\ \hline
Breast Cancer    & 569    & 30   & 2   & Image description\\ \hline
Pollen           & 301& 25000& 11& Single-cell RNA-seq \\ \hline
Levine 32        & 100000& 24& 14& Flow Cytometry \\ \hline
OrganAMNIST      & 8000& 784& 11& Medical Images \\ \hline
\end{tabular}%
%}

%\begin{flushleft}
\label{tab:datasetinformation}
%\end{flushleft}
\end{table}

\subsection{Competing Methods}

We compared VillageNet with six widely used clustering algorithms: K-Means, HDBSCAN, OPTICS and Agglomerative Clustering, Phenograph and the Scanpy implementation of Louvain clustering. These algorithms were selected on the basis of their wide usage and availability. K-Means and Agglomerative Clustering were provided with the true number of clusters for a fair comparison. HDBScan, OPTICS, and Agglomerative Clustering were excluded from testing MNIST and FMNIST datasets due to non-convergence within three hours.  

For biomedical datasets, we compared VillageNet against two clustering methods commonly used in biological data analysis: Phenograph and the Scanpy implementation of Louvain clustering. Phenograph is a nearest-neighbor graph–based method widely applied in flow-cytometry studies for its ability to capture complex, non-linear data structures. The Scanpy Louvain approach constructs a k-nearest-neighbor graph from the data and applies the Louvain community detection algorithm; it is extensively used in single-cell transcriptomic and other single-cell omics pipelines due to its scalability and its ability to capture complex, non-linear data structure.

\subsection{Comparison Metrics}

To compare the clustering results against known ground truth, we evaluate VillageNet using three performance measures: (i) Normalized Mutual Information (NMI) \cite{NMI1, NMI2} , (ii) Adjusted Rand Index (ARI) \cite{ARI}, and (iii) the number of clusters inferred by each algorithm. In addition, we report the wall-clock runtime to assess computational scalability.

NMI measures the amount of information shared between the predicted clustering and the ground-truth labels. Its values range from 0 to 1, where 0 indicates that there is no agreement beyond chance, and 1 corresponds to perfect agreement between the two partitions. In contrast, ARI quantifies agreement based on pairwise relationships between data points while correcting for agreement expected by chance. Similar to NMI, ARI values range from 0 to 1 in practice, with 0 indicating random assignment and 1 denoting perfect correspondence with the ground truth. Importantly, both NMI and ARI remain meaningful even when the number of inferred clusters differs from the true number of classes. Using both metrics rather than relying on a single measure provides a more balanced and reliable comparison between clustering algorithms.

We also report the number of clusters inferred by each algorithm. For K-Means and Agglomerative Clustering, the number of clusters was provided as an input rather than inferred automatically; these methods were nevertheless included for comparison using NMI and ARI. Wall-clock runtime is reported for all methods, and algorithms that failed to complete within three hours were excluded from the corresponding datasets.

\subsection{Hyperparameter Selection for VillageNet}

We made an initial heuristic choice for the hyperparameters: the number of villages was set to  $\nu = \min(N/10, 500)$, and the number of nearest neighbors was set to $\eta = 30$. This ensures that the average number of data points per village is roughly 10, which is substantially larger than a single point per cluster — the latter would render the k-means step ineffective.  Setting the number of villages to approximately $N/10$ also provides sufficient granularity for the linear approximations underlying the method. The upper limit of 500 prevents excessive computation time on very large datasets. Choosing $\eta = 30$ ensures that the connectivity of each village is based on a meaningful local structure rather than noise, while remaining small enough to avoid connections that are too broad and less informative.

Although these heuristics generally provide reasonable initial settings, they may not be optimal for every dataset. To address this, we developed a method to automatically select the hyperparameters without requiring knowledge of the ground truth. To identify optimal values, we assessed clustering stability across a two-dimensional grid of the two hyper-parameter settings.  For each pair of hyperparameters $(\nu, \eta)$, let $U(\nu, \eta)$ denote the resulting clustering.  We compute pairwise NMI scores between $U(\nu, \eta)$ and the clustering obtained from its four immediate neighboring parameter settings on the grid. The \textit{cluster stability score} for a given pair is defined as the mean of these four NMI values. Optimal hyperparameters are selected as those that maximize this score, ensuring robust and consistent clustering performance.  
For benchmarking VillageNet on biomedical datasets, where reproducibility and stability are critical, we used only optimal hyperparameter settings.

\subsection{Validation on Benchmark Datasets}

To validate the performance and general behavior of VillageNet, we first benchmarked it against a diverse collection of publicly available datasets with known ground-truth labels. Table \ref{tab:allexeperiments} summarizes the comparative performance of VillageNet against other clustering algorithms. Under both heuristic and optimized hyperparameter configurations, VillageNet consistently demonstrated strong and stable performance across datasets with varying dimensionality and noise characteristics.

VillageNet outperformed all other methods on UCI-HAR, Gas Sensor, Dry Beans, and FMNIST, and achieved results nearly identical to the best-performing algorithms on WiFi and Wine datasets. Even against Digits and MNIST, where VillageNet did not achieve the highest score, it performed comparably, achieving NMI values of 0.84 and 0.82, respectively, while also discovering 9 clusters, which is closer to the ground-truth value of 10 than that computed by competing approaches. This indicates that its inferred structure closely mirrors the true class organization, even without exact label correspondence.

Another key strength of VillageNet lies in its computational scalability and flexibility. As shown in Table \ref{tab:allexeperiments}, a single run of VillageNet (VillageNet-Heur) scales efficiently with the size of the data set, maintaining reasonable runtimes even for large and high-dimensional datasets such as MNIST and FMNIST. This efficiency enables VillageNet to perform rapid hyperparameter exploration and identify optimal configurations (VillageNet-Opt) without incurring significant computational cost. %Consequently, VillageNet provides a balanced combination of scalability, efficiency, and accuracy, which makes it particularly suitable for exploratory analysis of large and heterogeneous datasets.

%In general, these benchmark results confirm that VillageNet reliably captures the underlying data structure and determines an appropriate number of clusters ($m$) without prior knowledge. 
This validation establishes its robustness and scalability across heterogeneous data types—serving as a basis for subsequent evaluations on biomedical datasets.

\begin{table}[!ht] 
    \centering
    \caption{\bf Comparative Analysis of Unsupervised Clustering Algorithms on Publicly Available Datasets}
        \includegraphics[width=1\linewidth]{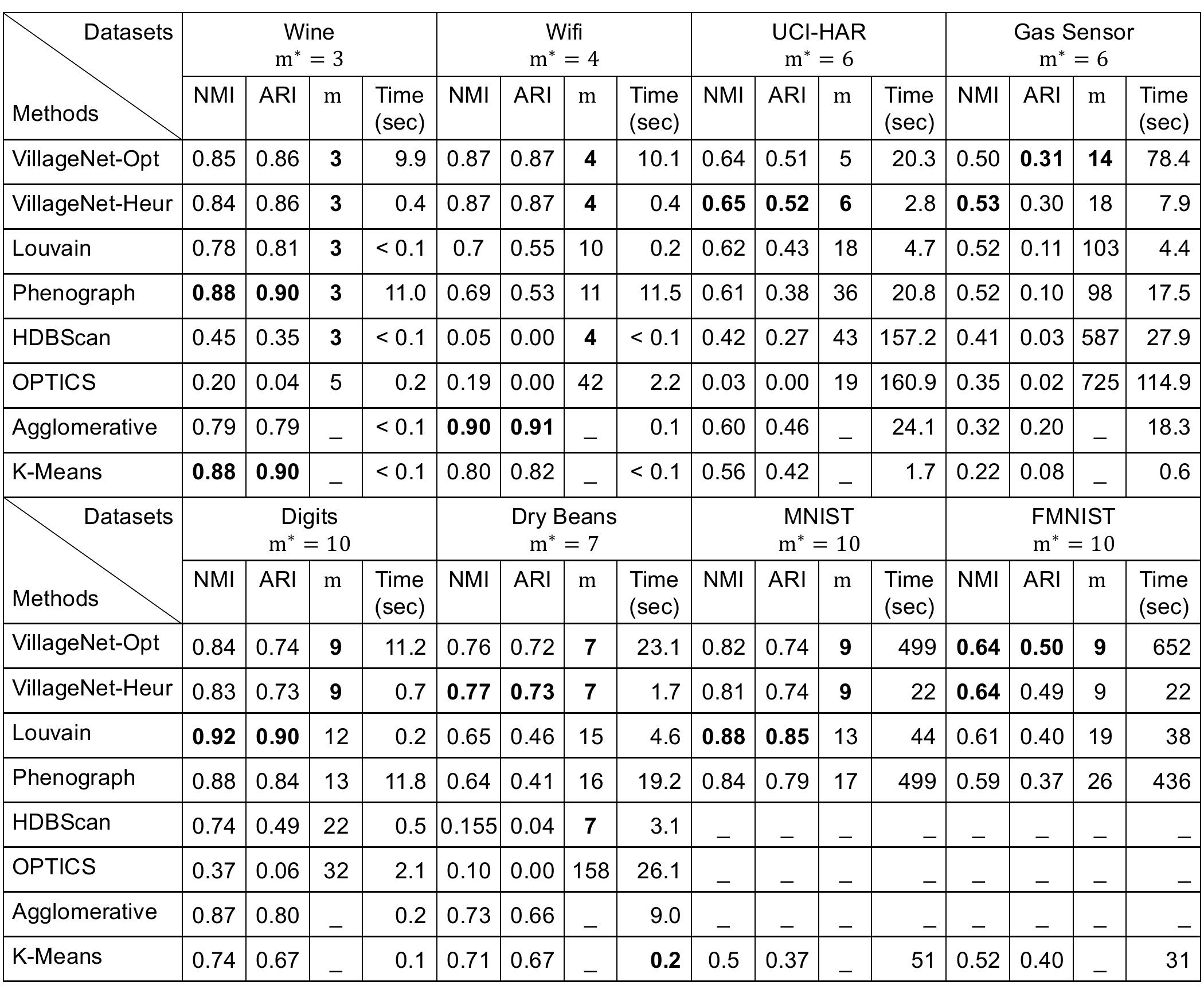}
        
    \label{tab:allexeperiments}
\end{table}

\subsection{Performance on Biomedical Datasets}

VillageNet was evaluated against four biomedical datasets with varying characteristics and complexities (Table \ref{tab:bioexperiments}). The Levine 32 dataset is a flow cytometry dataset with over 100,000 observations, frequently used as a reference for clustering methods. OrganAMNIST contains medical images of different organs, presenting challenges related to spatial variability and heterogeneity. The Breast Cancer dataset combines features derived from histopathological images and clinical data, while the Pollen dataset is a single-cell RNA sequencing data set (scRNA-seq) with $> 20,000$ genes from $N > 300$ cells, illustrating a high-dimensional data setting. We compared VillageNet performance with Phenograph and Scanpy Louvain, two widely used domain-specific clustering algorithms commonly applied in biomedical research. 
\begin{table}[!ht]
    \centering
    \caption{\bf Comparative analysis of VillageNet and competitive algorithms on biomedical datasets}
    \begin{tabular}{|c|c|c|c|c|c|c|c|c|c|l|l|l|}\hline
         &  \multicolumn{3}{|c|}{Exp1: Pollen}&  \multicolumn{3}{|c|}{Exp2: Levine 32}&  \multicolumn{3}{|c|}{Exp3: Breast Cancer} & \multicolumn{3}{|c|}{Exp4: ORGANAMNIST}\\\hline
 & NMI& ARI& m& NMI& ARI& m& NMI& ARI& m& NMI& ARI&m\\
 & & & (11)& & & (14)& & & (2)& & &(11)\\\hline
         VillageNet&  0.89&  0.76&  7&  \bf{0.95}&  \bf{0.97}&  8&  \bf{0.53}&  \bf{0.54}&  \bf{3}& \bf{0.84}& \bf{0.72}&\bf{8}\\
         Phenograph&  0.86&  0.78&  7&  0.81&  0.60&  24&  0.39&  0.21&  8& 0.69& 0.34&35\\
          Louvain&  \bf{0.92}&  \bf{0.84}&  \bf{8}&  0.86&  0.76&  \bf{17}&  0.38&  0.22&  7& 0.73& 0.39&38\\ \hline
    \end{tabular}
    \label{tab:bioexperiments}
\end{table}
Using the stability-based hyperparameter tuning method described in Section~4.2, VillageNet generated clusters that aligned closely with expert annotations in all four experiments. On \textit{Levine 32}, it achieved higher NMI and ARI scores than both Phenograph and Louvain. Although it identified eight clusters compared to fourteen in the ground truth, each identified cluster captured a biologically meaningful cell population. Specifically, as detailed in Table \ref{tab:FC}, $C_0$-$C_3$ corresponded to Monocytes, CD8 T cells, CD4 T cells, and Mature B cells, respectively. $C_4$ grouped Plasma B cells with Pre-B cells, $C_5$ grouped CD16$^{-}$ and CD16$^{+}$ NK cells, $C_6$ grouped CD34$^{+}$CD38$^{+}$CD123$^{-}$ HSPCs with CD34$^{+}$CD38$^{lo}$ HSCs and Pro-B cells, and $C_7$ grouped Basophils, CD34$^{+}$CD38$^{+}$CD123$^{+}$ HSPCs, and pDCs.  

\begin{table}[!ht]
    \centering
    \caption{\bf Comparison of VillageNet clusters with the ground truth on Levine32 dataset}
    \includegraphics[width=1\linewidth]{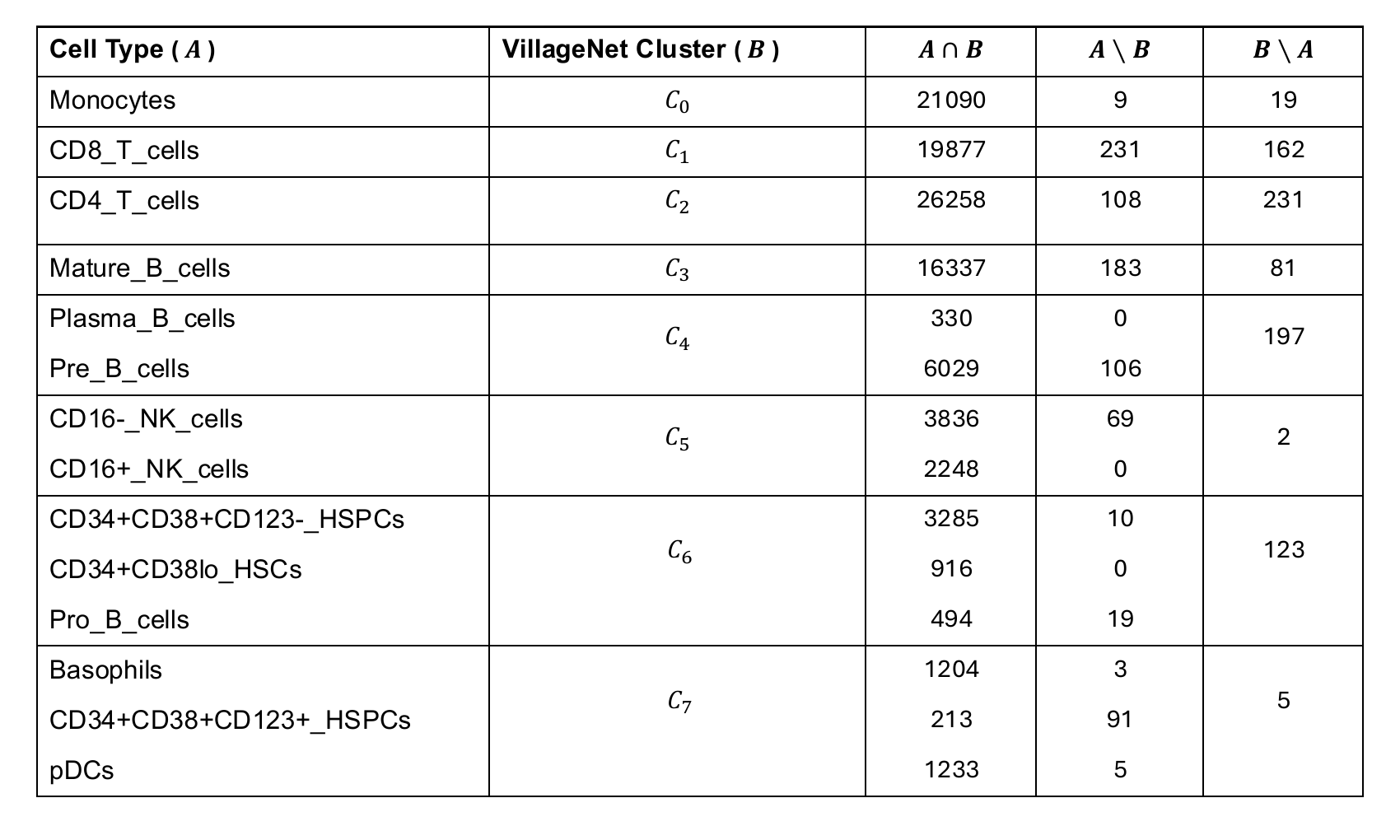}
    \label{tab:FC}
\end{table}
On \textit{OrganAMNIST}, VillageNet outperformed competing methods, successfully capturing complex spatial patterns in the imaging data. In the \textit{Breast Cancer} dataset, it achieved the highest accuracy metrics, underscoring its ability to integrate heterogeneous biomedical features.  

On the \textit{Pollen} scRNA-seq dataset, VillageNet’s NMI and ARI scores were slightly lower than those of Phenograph and Louvain but still remained high and biologically significant. The identified clusters reflected meaningful transcriptomic structure, with cluster–cell-type relationships reported in Table \ref{tab:SS}: $C_0$-$C_3$ corresponded directly to distinct cell types, $C_4$ largely matched the K562 population, $C_5$ grouped CRL-2339 with HL60, and $C_6$ captured multiple fetal cortex subpopulations. These results demonstrate that even in a high-dimensional and noisy environment, VillageNet revealed a coherent biological organization and produced interpretable clusters.  
\begin{table}[!ht]
    \centering
    \caption{\bf Comparison of VillageNet clusters with the ground truth on Pollen dataset}
    \includegraphics[width=1\linewidth]{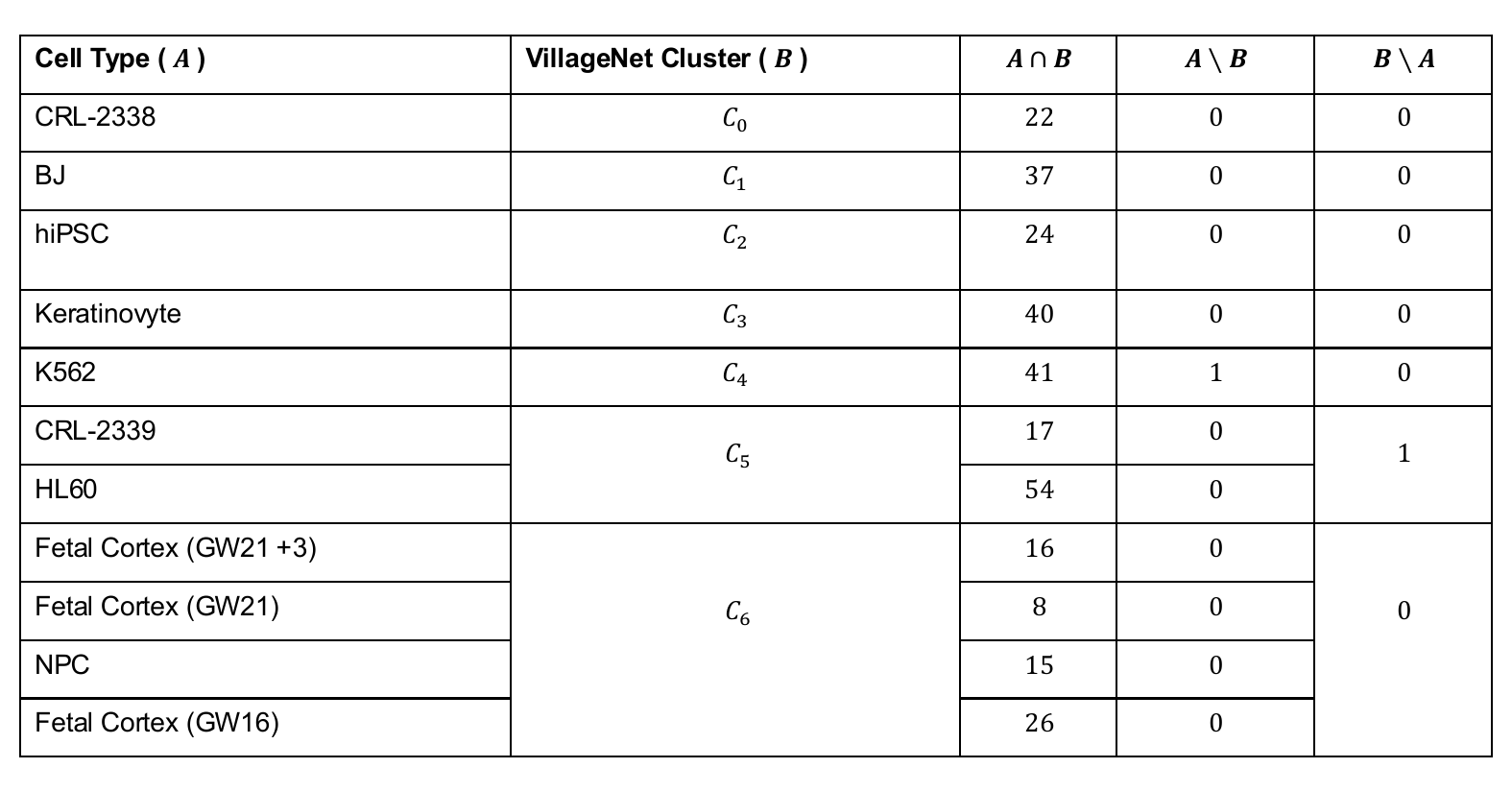}
    
    \label{tab:SS}
\end{table}

Taken together, these evaluations show that VillageNet balances clustering quality with adaptability across diverse biomedical data types---including flow cytometry, medical imaging, and transcriptomic. Its strong and consistent performance across datasets of different scales and modalities positions it as a flexible and broadly applicable clustering method for biomedical research.  

\section{Discussion}

Unsupervised clustering is commonly defined as the task of grouping datapoints such that instances within the same group are more similar to each other than to those in other groups, without access to training labels. Although intuitive, this definition does not correspond to a single rigorous mathematical objective. Instead, each clustering algorithm encodes its own notion of similarity and data structure and optimizes a corresponding objective function. As a result, different methods tend to perform well on datasets that align with their underlying assumptions, making broad applicability across diverse data types challenging. In this work, we develop and analyze the performance of a single clustering approach across datasets that vary in shape, size, dimensionality, and structure, with an emphasis on its behavior across multiple biomedical application. 

In precision medicine applications, clustering is evaluated primarily by its ability to identify clinically meaningful and interpretable subgroups from complex, heterogeneous data. In this context, assessing clustering performance requires consideration of multiple criteria that capture different aspects of clustering behavior, rather than relying on a single metric alone.

We evaluate VillageNet using three performance measures: (i) Normalized Mutual Information (NMI) \cite{NMI1, NMI2}, (ii) Adjusted Rand Index (ARI) \cite{ARI}, and (iii) the number of clusters inferred. NMI quantifies the information shared between predicted clusters and ground-truth labels, normalized to account for differences in cluster size and entropy, thus capturing global agreement between partitions. ARI measures agreement between cluster assignments based on pairwise relationships while correcting for agreement expected by chance, emphasizing local consistency in how data points are grouped. Evaluating both metrics along with the inferred number of clusters provides a more complete view of clustering quality, as each highlights different aspects of agreement.

Across publicly available non-biological benchmark datasets, VillageNet achieved the highest NMI and ARI values on 4 out of 8 datasets and inferred the number of clusters closest to the ground truth on all 8 datasets. On biomedical datasets, VillageNet achieved the best NMI and ARI on 3 out of 4 datasets and inferred cluster counts closest to the ground truth on 2 out of 4 datasets. Importantly, on datasets where VillageNet was not the top-performing method, its results remained highly competitive, indicating consistent recovery of the underlying structure rather than reliance on dataset-specific assumptions. 

Beyond these quantitative comparisons, VillageNet exhibits several properties that contribute to its practical usefulness and interpretability.

\subsection{Applicability across datasets with nonlinear structure}

VillageNet performs robustly on datasets with nonlinear cluster separation. Image datasets such as MNIST, FMNIST, and Digits are widely understood to exhibit complex, nonlinear class boundaries arising from high-dimensional variation \cite{nonlinearimage1, nonlinearimage2}. On these benchmarks, VillageNet achieved the best NMI and ARI scores on FMNIST, remained highly competitive on MNIST and Digits, and also performed strongly on ORGANAMNIST. These results indicate that VillageNet can reliably identify clusters even when class boundaries are not linearly separable.

\subsection{Consistency and interpretability of the number of clusters}

A central component of VillageNet is the Walk-Likelihood Community Finder (WLCF) \cite{WLCF}, which serves as the second-stage graph-clustering method and automatically determines the number of clusters, removing the need to pre-specify $m$. WLCF follows a top-down strategy: it first bifurcates the village graph to identify broad community structure, then refines these divisions using the Walk-Likelihood Algorithm, and finally merges communities when such merging increases modularity. Although WLCF optimizes the same modularity objective as algorithms such as Louvain, it inherently favors fewer and more cohesive clusters when modularity values are comparable. This preference helps prevent excessive fragmentation that can lead to very large numbers of clusters and hinders interpretability. Tables \ref{tab:allexeperiments} and \ref{tab:bioexperiments} show that VillageNet is the most consistent in determining a number of clusters closest to the ground truth. The difference is particularly pronounced for datasets such as UCI-HAR, Gas Sensor, and ORGANAMNIST, where competing algorithms substantially overfragment the data. For example, on ORGANAMNIST, VillageNet identified 8 clusters compared to the ground-truth value of 11, whereas Phenograph and Louvain produced 35 and 38 clusters, respectively. On the Gas Sensor dataset, VillageNet found 14 clusters (true value: 8), while Louvain generated 98 clusters and OPTICS produced 725. In practice, the true number of clusters is rarely known in advance; in such settings, severe overfragmentation increases the risk of fitting to noise rather than to meaningful structure. VillageNet’s conservative and coherent cluster selection therefore provides a critical advantage for interpretability and robustness.

\subsection{Operation in high-dimensional spaces without dimension reduction}

VillageNet also avoids several pitfalls associated with the curse of dimensionality, which strongly affects many high-dimensional biomedical datasets—particularly single-cell gene expression data, where each gene is a feature, and datasets frequently contain $>10,000$ genes. Many clustering algorithms, including Scanpy Louvain and even our previous work MapperPlus, attempt to address this challenge through dimension reduction; however, such preprocessing can introduce substantial distortions. Methods such as PCA optimize directions of maximal variance, but these directions are not necessarily the most biologically relevant for separating clusters \cite{bioPCAfault1, bioPCAfault2}. As a result, critical features for distinguishing cellular subpopulations can be suppressed or completely lost. At the same time, nearest-neighbor graphs, as used by methods such as Scanpy Louvain or Phenograph, become unreliable in high dimensions: as dimensionality increases, distances between points concentrate, making true neighbors difficult to distinguish from spurious ones \cite{nnissue}. Small fluctuations in high-dimensional space can, therefore, generate incorrect edges, leading to unstable and misleading cluster structures. VillageNet avoids these issues by constructing a graph whose nodes are entire villages rather than individual points, thereby reducing the sensitivity to high-dimensional noise. This enables VillageNet to operate directly in the native feature space while preserving meaningful structural relationships, without requiring any dimension-reduction preprocessing. In particular, on the Pollen single-cell dataset with $d=25,000$ genes and $N=301$ cells, VillageNet operates entirely in native space and achieves high NMI and ARI values compared to ground truth, demonstrating its effectiveness even in extremely high-dimensional regimes.

\subsection{Scalability}

VillageNet scales linearly with both the number of samples and the dimensionality of the data. This behavior is evident on large datasets such as MNIST and FMNIST and is particularly relevant for biomedical applications such as flow cytometry, where experiments routinely generate 50,000–100,000 data point i.e. cells. Linear scaling ensures that VillageNet remains practical for routine analysis rather than requiring prohibitively long computation times for large datasets.

\subsection{Usability}

Finally, VillageNet depends on only two hyperparameters. Empirically, simple heuristic choices perform well across a wide range of datasets. For users seeking additional guidance, a systematic procedure for selecting optimal hyperparameters is also provided, making the method straightforward to apply in practice.

\subsection{Summary}

In summary, VillageNet performs robustly across a wide range of datasets, including non-biological benchmarks as well as biomedical data spanning flow cytometry, single-cell gene expression, and image-based modalities. Its consistent performance across datasets of varying shape, size, dimensionality, and structure demonstrates that the method is largely hypothesis-free and does not rely on strong assumptions about cluster geometry. By combining local coarse-graining with graph-based community detection, VillageNet remains scalable, avoids reliance on dimension-reduction preprocessing, and yields coherent and interpretable cluster solutions. Together, these properties make VillageNet a scalable, robust, and interpretable clustering framework that is particularly well suited for broad biomedical applications.

\section*{Acknowledgments}
This work has been supported by grants from the United States National Institutes of Health (NIH) R01HL149431(LTI, YC), R01HL90880 (LTI, YC), R01HL123526 (LTI, YC), R01HL141460 (LTI, YC), R01HL159993 (LTI, YC), and R35HL166575 (LTI, YC, AB), U01 HL160274 (HeartShare JEL, LTI); and grants from the American Heart Association Strategically Focused Research Network (AHA SFRN) 23SFRNPCS1064232 (JEL, LTI, AB), 23SFRNPCS1060482 (LTI), and 23SFRNPCS1061606 (LTI).

Sandia National Laboratories (ED) is a multi-mission laboratory managed and operated by National Technology and Engineering Solutions of Sandia, LLC., a wholly owned subsidiary of Honeywell International, Inc., for the U.S. Department of Energy's National Nuclear Security Administration under contract DE-NA-0003525. This paper describes objective technical results and analysis. Any subjective views or opinions that might be expressed in the paper do not necessarily represent the views of the U.S. Department of Energy or the United States Government.

We are grateful for the unrestricted access to publicly available data utilized in this study.

\bibliography{bibliography}

\end{document}